\documentclass[letterpaper, 10 pt, journal, twoside]{IEEEtran}

\usepackage{amsmath}
\usepackage{graphicx}
\usepackage[colorlinks=true, allcolors=blue]{hyperref}
\usepackage{caption}
\usepackage{subcaption}
\usepackage{balance} 
\usepackage{ulem} 

\begin{document}
%
\title{HuNavSim: A ROS 2 Human Navigation Simulator for Benchmarking Human-Aware Robot Navigation}


\author{No\'e P\'erez-Higueras$^{1}$ and Roberto Otero$^{1}$ and Fernando Caballero$^{1}$ and Luis Merino$^{1}$
\thanks{Manuscript received: May, 4, 2023; Revised July, 15, 2023; Accepted September, 4, 2023.}
\thanks{This paper was recommended for publication by Editor Aniket Bera upon evaluation of the Associate Editor and Reviewers' comments.} 
\thanks{This work was partially supported by Programa Operativo FEDER Andalucia 2014-2020, Consejeria de Economía, Conocimiento y Universidades (DeepBot, PY20\_00817) and the projects NHoA (PLEC2021-007868) and NORDIC (TED2021-132476B-I00), funded by MCIN/AEI/10.13039/501100011033 and the European Union NextGenerationEU/PRTR.} 
\thanks{$^{1}$No\'e P\'erez-Higueras, Roberto Otero, Fernando Caballero and Luis Merino are with School of Engineering, Pablo de Olavide University, Crta. Utrera km 1, Seville, Spain
        {\tt\footnotesize noeperez@upo.es, rotegal@alu.upo.es, 
        fcaballero@upo.es, lmercab@upo.es}}%
\thanks{Digital Object Identifier (DOI): see top of this page.}
}

\markboth{IEEE Robotics and Automation Letters. Preprint Version. Accepted September, 2023}
{Pérez-Higueras \MakeLowercase{\textit{et al.}}: HuNavSim}

\maketitle

\begin{abstract}

This work presents the Human Navigation Simulator (\textit{HuNavSim}), a novel open-source tool for the simulation of different human-agent navigation behaviors in scenarios with mobile robots. The tool, the first programmed under the ROS 2 framework, can be used together with different well-known robotics simulators like Gazebo. The main goal is to facilitate the development and evaluation of human-aware robot navigation systems in simulation. In addition to a general human-navigation model, \textit{HuNavSim} includes, as a novelty, a rich set of individual and varied human navigation behaviors and an comprehensive set of metrics for social navigation benchmarking. 

\end{abstract}


\IEEEpeerreviewmaketitle

\section{Introduction}

\IEEEPARstart{T}{he} evaluation of the robot skills to navigate in scenarios shared with humans is essential to envisage new service robots working along people. The development of such mobile social robots poses two main challenges: first, real experimentation with people is costly, as well as being difficult to perform and to replicate except for very limited, controlled scenarios. In addition, human participants may be at risk, mainly during the initial stages of development. Therefore, simulating realistic human navigation behaviors for the development of robot navigation techniques is necessary. And secondly, the evaluation not only requires the assessment of the navigation efficiency, but also the safety and comfort of the people. This latter requirement is a human feeling which is difficult to quantify through mathematical equations. Thus, there is not solid agreement in the research community on a suitable set of metrics for human-aware navigation.   

Most state-of-the-art simulation approaches are based on models of crowd movement to control the behavior of simulated human agents. Whereas this is valid to obtain a collective behavior of the agents, it loses realism at the local level since the behavior of all individual agents is exactly the same for the same scenarios. Here, we propose a set of individual human behaviors related to reactions to the presence of a robot. 

Another issue is the evaluation of the human-aware navigation through metrics. Each benchmarking tool usually presents its own set of metrics. While research in new realistic ``social'' metrics is needed, the absence of common well-known metrics hinders the comparison of different techniques for social robot navigation.

\begin{figure}[!t] 
    \centering
    \includegraphics[scale=0.25]{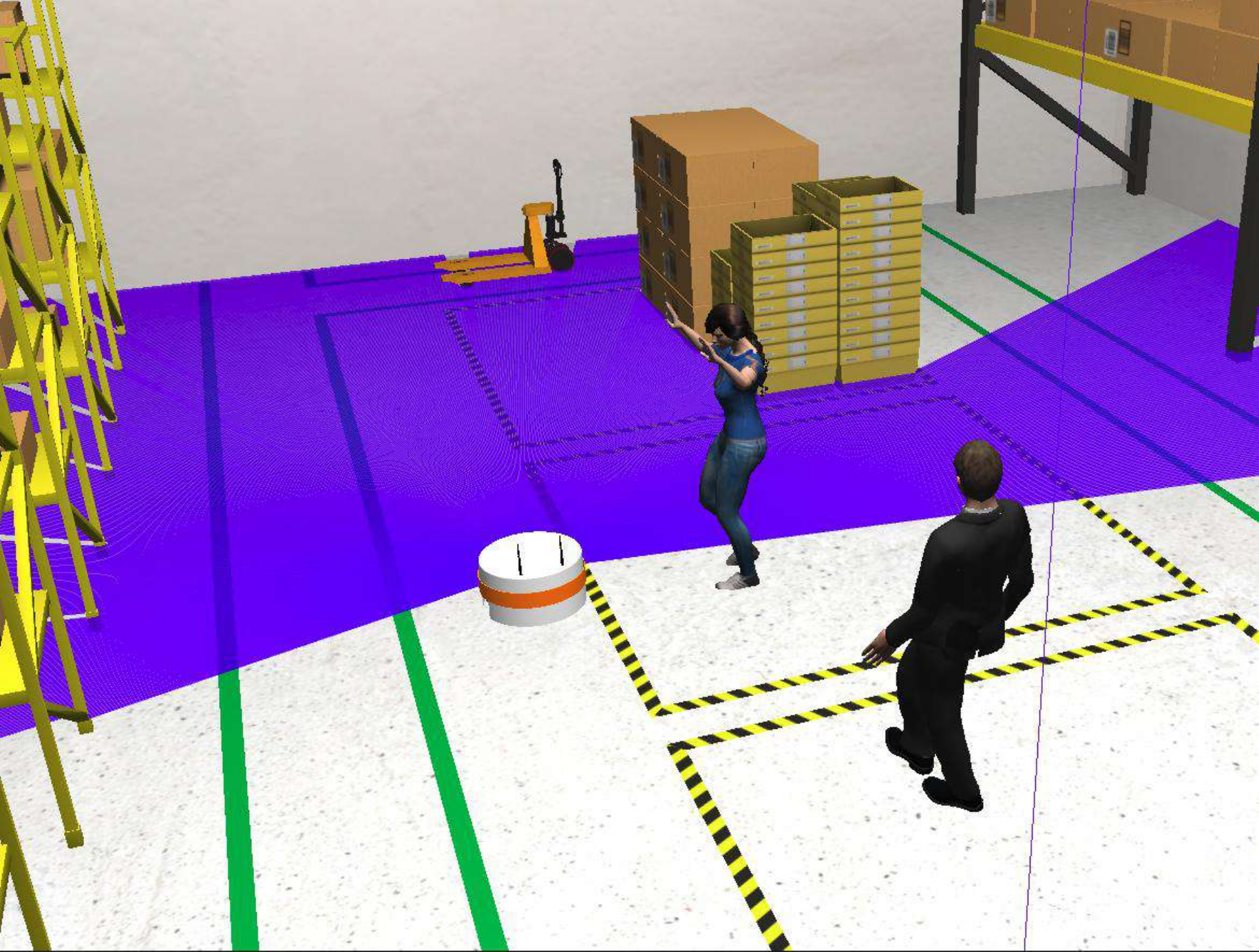}
    \caption{Capture of HuNav agents in the Gazebo Simulator}
    \label{fig:gazebo_people}
\end{figure}

With the \textit{HuNavSim} we aim to contribute to the solutions of the two problems mentioned: providing a set of different behaviors for individual agents closer to the reality, and presenting a compilation of metrics employed in the literature. In a nutshell, we present the following contributions: 

\begin{itemize}
    \item[i)] An open-source and flexible simulation tool of human navigation under the ROS 2 framework \cite{ros2} that can be used together with different robotics simulators.
    \item[ii)] A rich set of navigation behaviors of the human agents, which includes a set of frequent individual reactions to the presence of a robot.
    \item[iii)] A wide compilation of metrics into the tool from the literature for the evaluation of human-aware navigation, which is configurable and extensible.
    \item[iv)] A wrapper to use the tool along with the well-known Gazebo simulator used in Robotics (see Fig. \ref{fig:gazebo_people}). 
\end{itemize}

\section{Related Work}

\begin{figure*}[!t] 
      \centering
      \includegraphics[scale=0.63]{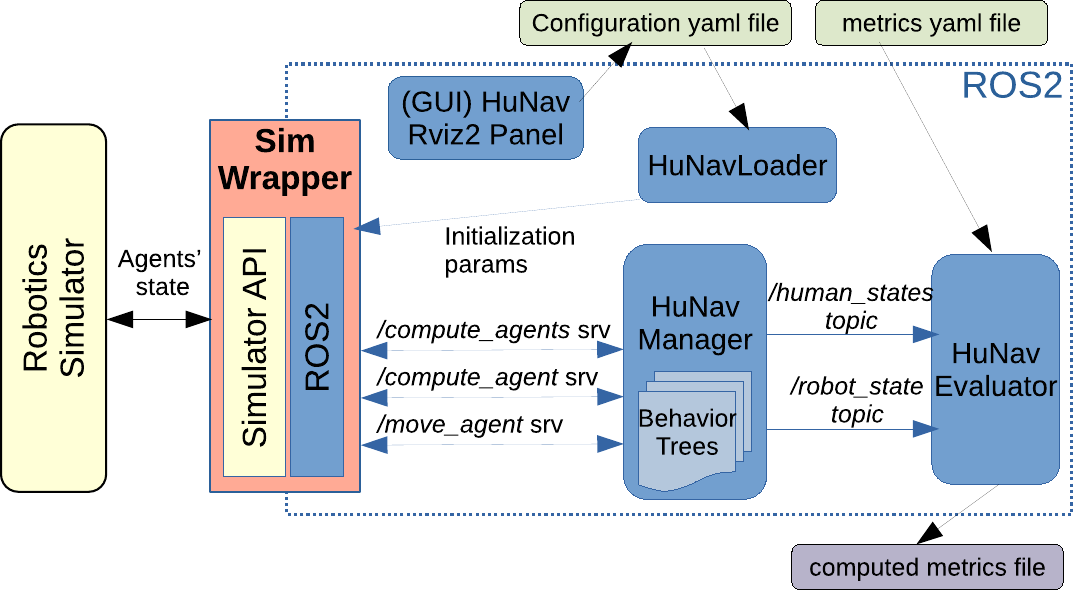}
      \caption{Diagram of \textit{HuNavSim}. The core modules of \textit{HuNavSim} are in blue. The base robotics simulator modules are depicted in light yellow color. Green boxes indicate the configuration input data. Output evaluation metrics are indicated in light purple color.}
      \label{fig:diagram}
\end{figure*}

Different simulators and benchmarking tools for human-aware navigation problems can be found in the literature. We provide a brief review of existing related software and highlight the differences and similarities with our approach.  

\textit{PedSimROS}\footnote{\url{https://github.com/srl-freiburg/pedsim_ros}} is an adaptation to ROS of a basic library that uses Social Force Model (\textit{SFM}) \cite{Helbing1995SocialFM} to lead the crowd movement. \textit{MengeROS}\footnote{\url{https://github.com/ml-lab-cuny/menge_ros}} \cite{MengeROS_Aroor2017} is a more comprehensive tool that uses $A^*$ and potential fields 
for global path planning, and allows for the selection among a set of collision-avoidance strategies such as different variations based on SFM, Optimal Reciprocal Collision Avoidance (ORCA) \cite{orca_2011} and Pedestrian Velocity Obstacle (PedVO) \cite{PedVO_2012}. However, both PedsimROS and MengeROS are deprecated software integrated into unmaintained versions of ROS 1. Also, they do not incorporate any option for navigation evaluation.
\textit{CrowdBot}\footnote{\url{http://crowdbot.eu/CrowdBot-challenge/}} \cite{crowdbot_ICRA21} and \textit{ SEAN (Social Environment for Autonomous Navigation)} \footnote{\url{https://sean.interactive-machines.com/}} \cite{sean2_Tsoi_RAL22, sean_Tsoi20} are more recent, advanced and ambitious tools. They share similar features. Both are based on the game engine Unity and ROS 1, and both aim at becoming the standard for evaluating robot navigation in populated environments. In contrast to them, \textit{HuNavSim} provides a more flexible approach that allows using the tool along with different simulators and provides a set of individual and more realistic human reactions to the presence of a robot. Moreover, these tools present a closed set of metrics while \textit{HuNavSim} includes a larger compilation of metrics that is easily configurable and extendable.     

Another interesting simulator is the \textit{Intelligent Human Simulator (InHuS)} \cite{inhus_favier21a, inhus_favier21b}. This simulator is meant to control the movement of human agents in another simulator. It also includes a small set of human individual behaviors, as \textit{HuNavSim} does. In contrast to \textit{InHuS}, our simulator employs a powerful and flexible tool, behavior trees \cite{behavior_trees}, to define and drive particular human behaviors. These trees can also be easily modified and extended. Moreover, \textit{InHuS} is based on ROS 1 and it mainly employs a navigation system devised for robots, \textit{HATEB2} \cite{hateb2_teja20}, to control human movements. This could lead to more unrealistic crowd movement than simulators based on specific crowd movement models.

\textit{SocNavBench}\footnote{\url{https://github.com/CMU-TBD/SocNavBench}} \cite{socnavbench_Biswas21} follows a different approach. It is a simulator-based benchmark with pre-recorded real-world pedestrian data replayed. The main drawback of this tool is that pedestrians' trajectories are replayed from open-source datasets and, therefore, the effects of robot motions in pedestrians' paths are not considered. That makes it difficult to obtain a realistic evaluation of the human-aware navigation. 

Other approaches are devised for training, testing, and benchmarking learning-based navigation agents in dynamic environments. For instance, \textit{Arena-Rosnav} \cite{arena_kastner2021} allow for the training of obstacle avoidance approaches based on Deep Reinforcement Learning.
It makes use of \textit{Pedsim} library to guide pedestrians' movements. Then, \textit{Arena-Rosnav} was extended to include navigation benchmarking,  \textit{Arena-Bench} \cite{arena-bench}, and a larger set of utilities for development and benchmarking  (\textit{Arena-rosnav 2.0} \cite{arena-rosnav2023}). For the simulation and training of reinforcement learning agents in spaces shared with humans, we found \textit{SocialGym 2.0} \cite{holtz2022socialgym, socialgym2023}, which also uses the \textit{Pedsim} library to simulate human navigation. These tools do not include particular human behaviors nor a specific set of metrics for human-aware navigation.       

Finally, there are interesting evaluation tools like \textit{BARN (Benchmark for Autonomous Robot Navigation)}\footnote{\url{https://www.cs.utexas.edu/~attruong/metrics_dataset.html}} \cite{BARN_Perille2020} and \textit{Bench-MR}\footnote{\url{https://github.com/robot-motion/bench-mr}} \cite{bench-mr_21}. However, those are oriented to the general problem of navigation in cluttered environments without considering the ``social'' component of spaces shared with humans.

\section{Human Navigation Simulator (HuNavSim)}

\subsection{Architecture of the simulator}

The general architecture of the simulator can be seen in Fig. \ref{fig:diagram}. \textit{HuNavSim} is in charge of properly controlling the pose in the scene of the human agents spawned in another base simulator such as Gazebo, Morse, or Webots, which also must simulate the scenario and the robot. Therefore, a wrapper to communicate with the base simulator is required. 

Initially, the number and characteristics of the agents to be simulated (like individual behavior, list of goals, etc) must be provided by the user. It can be specified through a configuration \texttt{yaml} file, or through a graphic user interface based on a ROS 2 RViz panel.

Then, at each execution step, the simulator wrapper sends the current human agents' status to the \textit{hunav\_manager} module through ROS 2 service calls. According to the current state, the system decides the next state of the agents which are returned to the wrapper, and thus updated in the base simulator.  

Finally, the \textit{hunav\_evaluator} module records the data of the experiment and computes the evaluation metrics at the end of the simulation. The information about each simulation and the desired set of metrics to be computed can be specified through a \texttt{yaml} file or through a list of checkboxes shown in a ROS 2 RViz panel. The module generates a set of output result files with the simulation info and the names and values of the computed metrics.  

The complete documentation and code of the \textit{HuNavSim} is available at \url{https://github.com/robotics-upo/hunav_sim}

\subsection{Human navigation modeling}

The \textit{HuNavSim} is primarily based on the use of the well-known Social Force Model \cite{Helbing1995SocialFM} and its extension for groups \cite{Mussaid_ext_sfm09, Mussaid_ext_sfm10}, to lead the human agents movement similar to other crowd simulators. The simplicity and good results of this model allowed us to easily extend it to provide a set of different and individual navigation reactions of the agents to the presence of the robot. That allows for the enrichment of the navigation scenarios and to challenge the navigation algorithms with more diverse and realistic human behaviors. The set of behaviors included are the following:  

\begin{itemize}
    \item \textit{regular}: this is the regular navigation based on SFM. The robot is included as another human agent, so the human treats the robot like another pedestrian.
    \item \textit{impassive}: in this case, the robot is not included as a human agent in the SFM. Therefore, the human deals with the robot as an obstacle.
    \item \textit{surprised}: when the human sees the robot, the current goal navigation is stopped. The human stops walking and starts to look at the robot (heading to the robot position).
    \item \textit{curious}: the human abandons his/her current navigation goal when the robot is detected. Then, the human approaches the robot slowly. After a short time ($30$ seconds by default) the human loses interest and goes back to the original navigation goal.   
    \item \textit{scared}: when the human agent detects the robot, it tries to stay far away from it. In this case, we add to the agent an extra repulsive force from the robot and we decrease the maximum velocity. This way we simulate a scared human behavior.   
    \item \textit{threatening}: when the agent sees the robot, it approaches the robot and tries to block the robot's path by walking in front of it. A goal in front of the robot is continuously computed and commanded. After some time ($40$ seconds by default), the agent gets bored and continues its original navigation.    
\end{itemize} 

An interesting feature is that all these behaviors are controlled by behavior trees \cite{behavior_trees}. They make possible efficient structuring of the switching between the different tasks or actions of the autonomous agents. Moreover, they are easily programmable, and new behaviors can be added or modified with minimal effort. We use the engine \textit{BehaviorTree.CPP}\footnote{\url{https://github.com/BehaviorTree/BehaviorTree.CPP}} for behavior creation and editing.    

\subsection{Wrapper for Gazebo Simulator}

   \begin{figure}[!t] 
      \centering
      \includegraphics[scale=0.47]{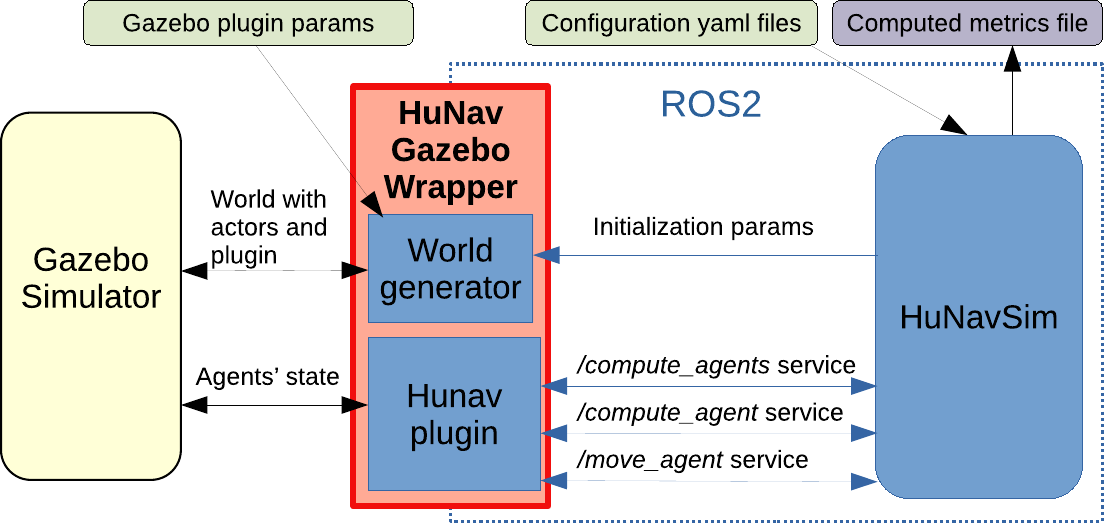}
      \caption{Diagram of the Gazebo wrapper for \textit{HuNav} Simulator. The red square encapsulates the two modules of the wrapper, which intermediates between the Gazebo simulator and the \textit{HuNavSim}.}
      \label{fig:wrapper_diagram}
   \end{figure}

A wrapper to use \textit{HuNavSim} along with the Gazebo simulator is also provided. Figure \ref{fig:wrapper_diagram} shows the wrapper modules (red square) and the communication with the Gazebo simulator and \textit{HuNavSim}. 

The control of the agent movement has been programmed as a Gazebo world plugin that must be included in the Gazebo world file. Also the human agents are included as a Gazebo actor in the same file.
For that reason, we have programmed the following two modules: 
\begin{itemize}
    \item A \textit{world\_generator} module which is in charge of reading the configuration parameters of the human agents (defined in the agents' configuration \texttt{yaml} file), and to write the proper plugin and agents in the base Gazebo world file.
    \item Then, Gazebo loads this world file and the \textit{HuNav plugin} communicates with \textit{HuNavSim} to update the agents' status (model pose in the world) during the execution of the simulation. 
\end{itemize}

A set of typical scenarios shared with humans is included in the wrapper: a cafeteria, a warehouse, and a house. Moreover, we have developed a new set of different 3D human models. The human 3D model and the associated behavior can be selected by the user. Since \textit{HuNavSim} only determines the agents' global position and orientation in the Gazebo world, we try to provide an agent's natural body movements by using different animations to better express the different agent behaviors visually. However, producing a very realistic body movement is beyond the scope of this work.     

The wrapper documentation and code are publicly available at \url{https://github.com/robotics-upo/hunav_gazebo_wrapper}

\begin{table*}[t]
\caption{Available metrics in \textit{HuNavSim}}
\label{tab:metrics_list}
\begin{center}
\begin{tabular}{|c|c|l|c|}
\hline 
Metric & Source & Description & units \\
\hline \hline

time\_to\_reach\_goal & Pérez et al.\cite{ijsr18irlrrt} & Time spent to reach the goal & $s$\\
\hline
path\_length & Pérez et al.\cite{ijsr18irlrrt} & Length of the robot path & $m$\\
\hline
cumulative\_heading\_changes & Pérez et al.\cite{ijsr18irlrrt} & Added absolute heading angle differences between successive waypoints & $rad$\\
\hline
avg\_dist\_to\_closest\_person & Pérez et al.\cite{ijsr18irlrrt} & Distance to the closest person at each time step on average & $m$\\
\hline
min\_dist\_to\_people & Pérez et al.\cite{ijsr18irlrrt} & The minimum distance to any person along the trajectory & $m$\\
\hline
intimate\_space\_intrusions & Pérez et al.\cite{ijsr18irlrrt} & Time percentage the robot spent in a person intimate space. See Proxemics\cite{proxemics_Hall} & $\%$\\
\hline
personal\_space\_intrusions & Pérez et al.\cite{ijsr18irlrrt} & Time percentage the robot spent in a person personal space \cite{proxemics_Hall} & $\%$\\
\hline
social+\_space\_intrusions & Pérez et al.\cite{ijsr18irlrrt} & Time percentage the robot spent in a person social or public space \cite{proxemics_Hall} & $\%$ \\
\hline
group\_intimate\_space\_intrusions & Pérez et al.\cite{ijsr18irlrrt} & Time percentage the robot spent in a group intimate space & $\%$\\
\hline
group\_personal\_space\_intrusions & Pérez et al.\cite{ijsr18irlrrt} & Time percentage the robot spent in a group personal space & $\%$\\
\hline
group\_social+\_space\_intrusions & Pérez et al.\cite{ijsr18irlrrt} & Time percentage the robot spent in a group social or public space & $\%$ \\
\hline

completed & \textit{SEAN2.0} \cite{sean2_Tsoi_RAL22} & Whether the robot reached the goal or not & - \\
\hline
min\_dist\_to\_target & \textit{SEAN2.0} \cite{sean2_Tsoi_RAL22} & The closest the robot passes to the target (or goal) position & $m$ \\
\hline
final\_dist\_to\_target & \textit{SEAN2.0} \cite{sean2_Tsoi_RAL22} & Distance between the last robot position and the target (goal) position & $m$\\
\hline
robot\_on\_person\_collisions & \textit{SEAN2.0} \cite{sean2_Tsoi_RAL22} & Number of times robot collides with a person & - \\
\hline
person\_on\_robot\_collisions & \textit{SEAN2.0} \cite{sean2_Tsoi_RAL22} & Number of times person collides with a robot & - \\
\hline
time\_not\_moving & \textit{SEAN2.0} \cite{sean2_Tsoi_RAL22} & Seconds that the robot was not moving & $s$\\
\hline

avg\_robot\_linear\_speed & \textit{SocNavBench}\cite{socnavbench_Biswas21} & Average linear speed of the robot along the trajectory & $m/s$\\
\hline
avg\_robot\_angular\_speed & \textit{SocNavBench}\cite{socnavbench_Biswas21} & Average absolute angular speed of the robot along the trajectory & $rad/s$\\
\hline
avg\_robot\_acceleration & \textit{SocNavBench}\cite{socnavbench_Biswas21} & Average acceleration of the robot along the trajectory & $m/s^2$\\
\hline
avg\_robot\_jerk & \textit{SocNavBench}\cite{socnavbench_Biswas21} & Average jerk (time derivative of acceleration) of the robot along the trajectory & $m/s^3$\\
\hline

avg\_pedestrians\_velocity & Katyal et al.\cite{KapilIROS22} & Average linear velocity of all the pedestrian along the trajectory & $m/s$\\
\hline
avg\_closest\_pedestrian\_velocity & Katyal et al.\cite{KapilIROS22} & Average linear velocity of the closest pedestrian along the trajectory & $m/s$\\
\hline

social\_force\_on\_agents & SFM-based\cite{Helbing1995SocialFM} & Sum of the modulus of the social forces provoked by the robot on the & $m/s^2$ \\ & & human agents along the trajectory & \\
\hline
social\_force\_on\_robot & SFM-based\cite{Helbing1995SocialFM} & Sum of the modulus of the social forces provoked by the agents on the & $m/s^2$ \\ & & robot along the trajectory & \\
\hline
obstacle\_force\_on\_agents & SFM-based\cite{Helbing1995SocialFM} & Sum of the modulus of the obstacle forces on the agents along the trajectory & $m/s^2$ \\
\hline
obstacle\_force\_on\_robot & SFM-based\cite{Helbing1995SocialFM} & Sum of the modulus of the obstacle forces on the robot along the trajectory & $m/s^2$ \\
\hline
social\_work & SFM-based\cite{Helbing1995SocialFM} & The sum of the social force on the robot, the obstacle force on the robot, & $m/s^2$ \\ & & and the social force on the agents & \\
\hline

\end{tabular}
\end{center}
\end{table*}

\section{Navigation evaluation}

With the aim of tackling the problem of selecting the best metrics for evaluation of the human-aware navigation, we decided to keep the evaluation system as open as possible.

First, we reviewed the literature in order to collect most of the metrics applied to the issue. Then, we let the user select the metrics to be computed for each simulation. Finally, the system also allows for the easy addition of new metrics to the evaluation. The input of all the metric functions consists of two arrays with the poses, velocities, and other data of the \textit{HuNavSim} agents and the robot for each time step.

Besides the computation of the metrics for all the agents in the scenario, the evaluator also performs the calculation regarding the particular behavior of the agents. In this way we can obtain interesting metrics for the situations in which the agents show a specific behavior to the presence of the robot.

The main objective here is to validate the usefulness of HuNavSim for comparing planners under the given model. The main contribution is the possibility of including individualized behaviors, which therefore allows one to evaluate the same planners in a wider range of situations.

Thus, \textit{HuNavSim} presents a reliable and flexible evaluation system in contrast to the fixed evaluation systems currently found in the literature. 

  \begin{figure*}[!t] 
      \centering
      \includegraphics[scale=0.310]{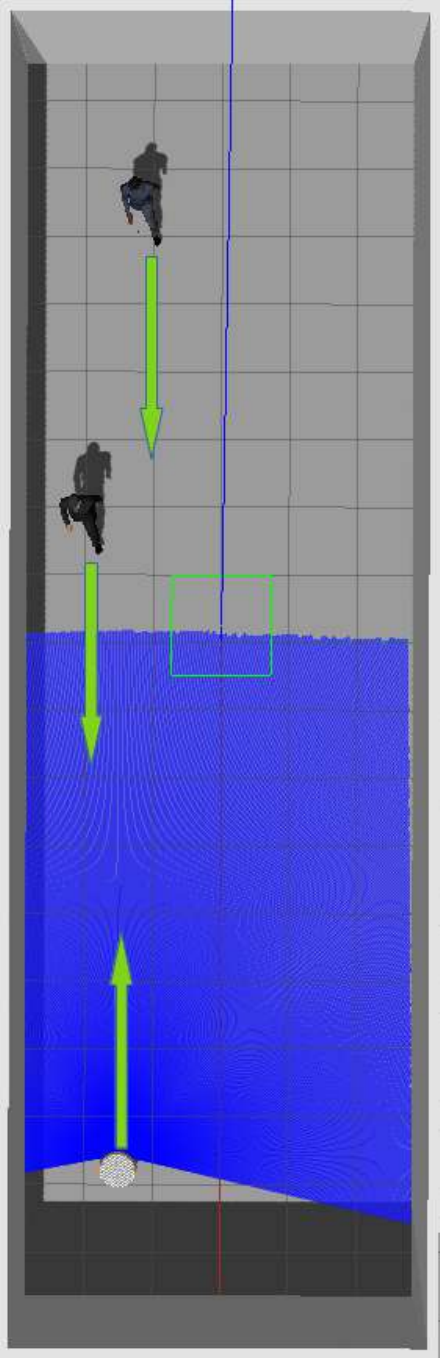}
      \includegraphics[scale=0.305]{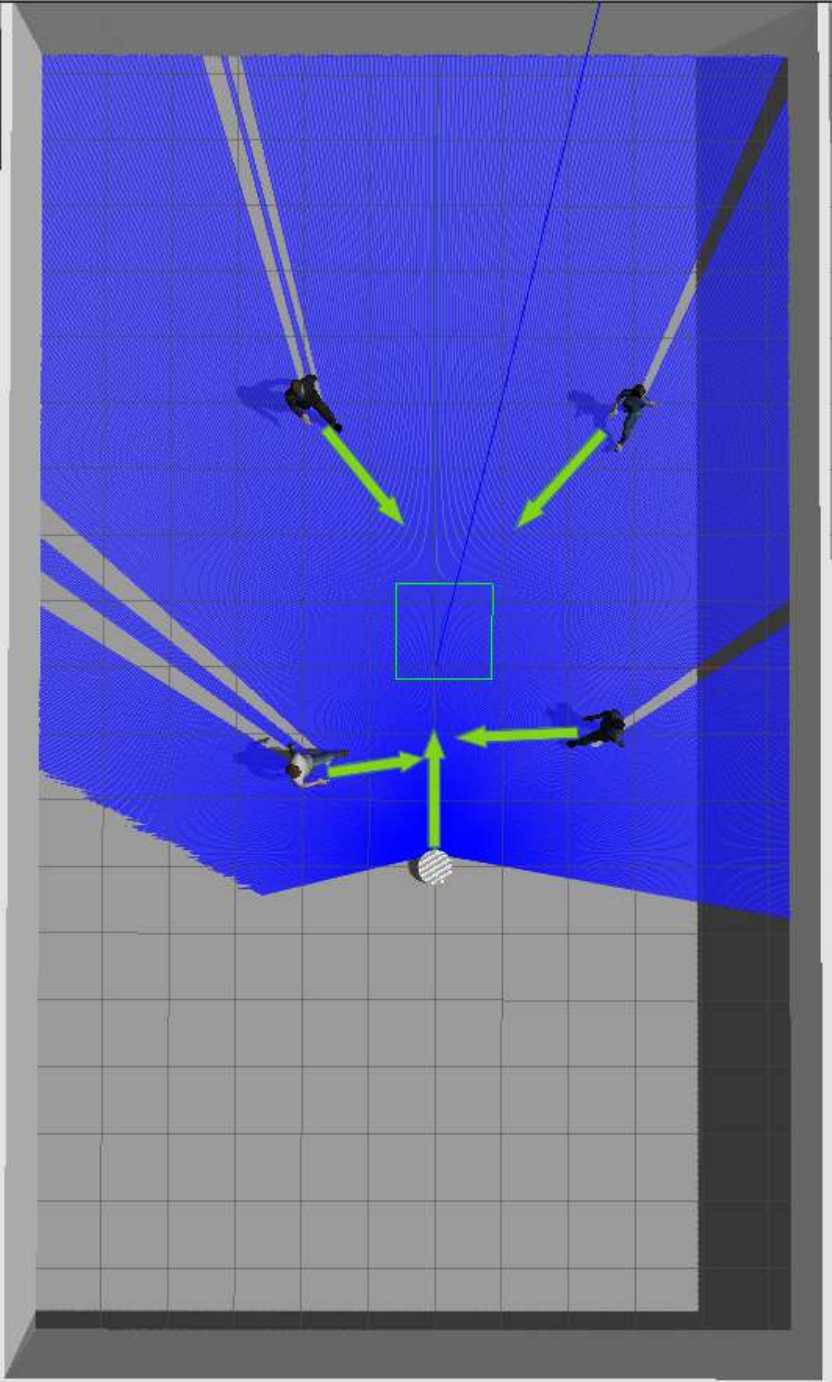}
      \includegraphics[scale=0.324]{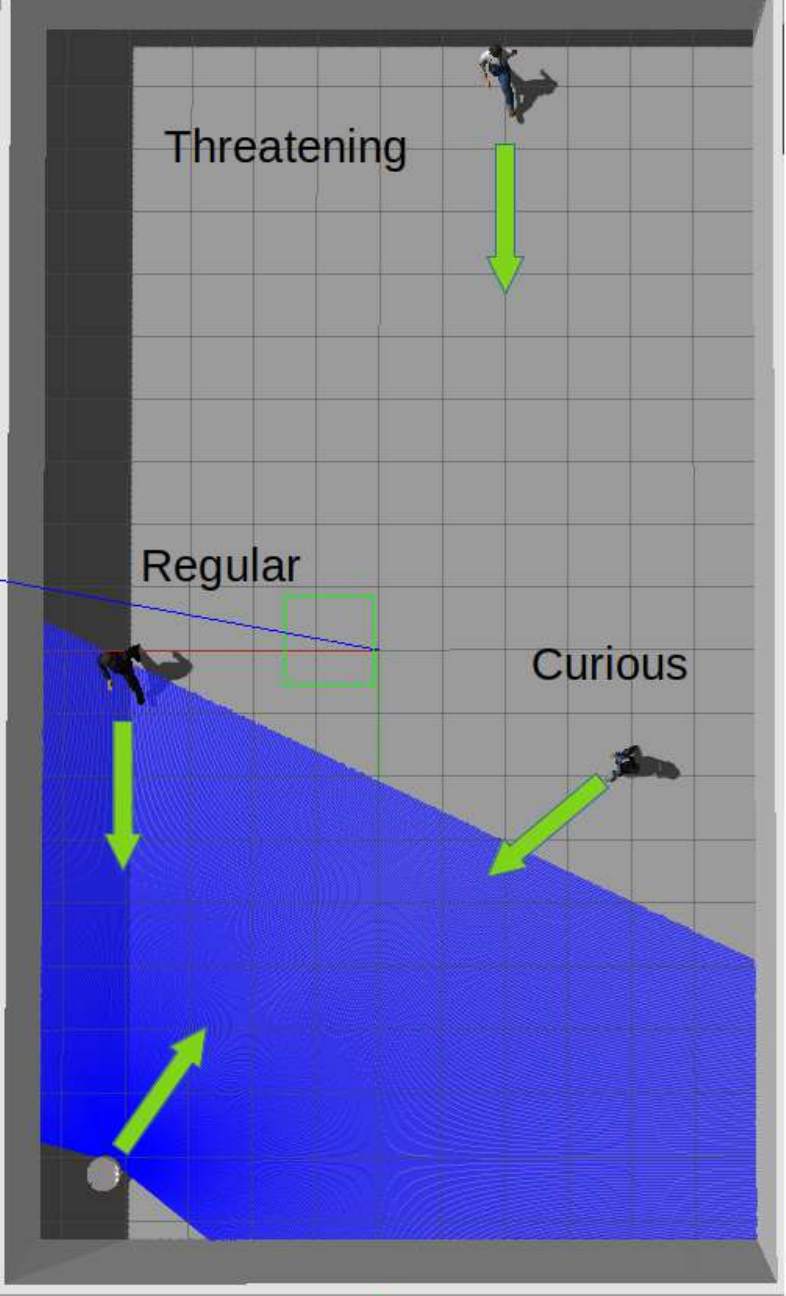}
      \caption{Scenarios for evaluation. Left image: passing scenario in a corridor, center image: crossing scenario, right image: combined scenario with different human navigation behaviors. Green arrows indicate the direction of the navigation goal for the agents and robot.}
      \label{fig:scenarios}
   \end{figure*}

\subsection{Metrics description}

At the time of writing this work, the metrics implemented were those employed in our previous work \cite{ijsr18irlrrt}, the \textit{SEAN} simulator \cite{sean2_Tsoi_RAL22}, some of the \textit{SocNavBench} \cite{socnavbench_Biswas21},  and some metrics related to the Social Force Model (SFM) employed in different works according to the compilation of Gao \emph{et al}. \cite{gao_frontiers22}. This results in a total of 28 metrics, and it represents, to the best of our knowledge, the most extensive compilation of metrics. The list of available metrics can be seen in Table \ref{tab:metrics_list}. A detailed description of the metrics along with their mathematical equations can be reviewed on the Github page of the \textit{HuNavEvaluator}\footnote{\url{https://github.com/robotics-upo/hunav_sim/tree/master/hunav_evaluator}}  


\subsection{Output metrics data}

As output, the HuNavEvaluator will generate a set of metrics files. They are formed as text files starting with headers indicating the metrics names, followed by the data organized in a tabular form of rows and columns:
\begin{itemize}
    \item General metrics file, with the values of the final metrics computed.
    \item List of metrics values computed for each time step in the simulation. This is very useful to plot and to visualize the behavior of some metrics over the simulation time or at specific moments. 
    \item Set of files with the final metrics but only considering the groups of people with the same behavior. We will obtain as many files as particular behaviors we use in the simulation. 
    \item Finally, the tool will generate the metrics computed at each time stamp for each previous behavioral group.
\end{itemize}

\section{Validation experiments}

To assess the \textit{HuNavSim} as a benchmarking tool, we employ three different planners in different socially-aware navigation situations. Then, the metrics produced by the tool are used to compare the social compliance of the navigation.

\subsection{Scenarios}

According to Gao \emph{et al}. \cite{gao_frontiers22}, the scenarios commonly used in evaluating human-aware navigation are: \texttt{Passing}, \texttt{Crossing}, \texttt{Overtaking}, \texttt{Approaching}, \texttt{Following/Accompanying} and a \texttt{Combination} of those.
In this evaluation, we employ the basic scenarios for \texttt{Passing} and \texttt{Crossing}, and a \texttt{Combined} scenario with humans showing different individual behaviors. In Fig. \ref{fig:scenarios} we can see the \texttt{Passing} scenario in the left image, in which the robot must reach the other side of the room while two people are walking in the opposite direction. In case of the \texttt{Crossing} scenario, center image of Fig. \ref{fig:scenarios}, the robot must again reach the opposite side of the room while crossing with four people: two of them walk perpendicularly in the robot's direction, and the other two walk diagonally. Finally, the right image presents the \texttt{Combined} scenario with three people showing different reactions to the presence of the robot: a ``regular'' navigation, which deals with the robot as a another human; a ``curious'' reaction, in which the human agent carefully approaches the robot; and a ``threatening'' reaction, in which the human agent will insistently try to block the robot from passing.

\subsection{Planners}

In the previous social-navigation scenarios, we will compare the metrics obtained by three planners with different features: 

\begin{itemize}
    \item \textit{DWB}. The basic ROS 2 navigation stack using the DWB local planner\footnote{\url{https://github.com/ros-planning/navigation2/tree/main/nav2_dwb_controller}} which is based primarily on the Dynamic Window Approach \cite{dwa_97} to reach a goal efficiently without ``social'' constrains.
    \item \textit{SCL}. The same system using a Social Costmap Layer (a custom ROS 2 version of the social navigation layers implemented in ROS 1\footnote{\url{https://github.com/robotics-upo/nav2_social_costmap_plugin}}). It includes a cost around the people according to the distance and orientation.
    \item \textit{SFW}. A social local planner, developed by us, based on the Dynamic Window Approach\footnote{\url{https://github.com/robotics-upo/social_force_window_planner}}. It uses the Social Force Model as a human pose predictor and a trajectory scoring function which adds the social work as a new term. That means that in addition to distance from people, the velocity vectors of the robot and people are also considered.
\end{itemize}

\subsection{Results}

\subsubsection{Results for regular navigation behavior}

First, we show the general metrics obtained by the planners in the \texttt{Passing} and \texttt{Crossing} scenarios. We have repeated the trajectory $10$ times for each planner in both scenarios. Tables \ref{tab:passing} and \ref{tab:crossing} show the average values of the metrics obtained. 

\begin{table}[t]
\caption{Metrics for the \texttt{Passing} scenario}
\label{tab:passing}
\begin{center}
\begin{tabular}{|c||c|c|c|c|}
\hline 
\texttt{Passing} & \textit{DWB} & \textit{SCL} & \textit{SFW} & units \\
\hline \hline
time\_to\_reach\_goal & \boldmath $25.5$ & $26.6$ & $35.7
$ & $s$\\
\hline
path\_length & \boldmath $15.2$ & $15.7$ & \boldmath $15.2$ & $m$\\
\hline
cumulative\_heading\_changes & \boldmath $0.47$ & $1.89$ & $1.89$ & $rad$\\
\hline
avg\_dist\_to\_closest\_person & $6.36$ & \boldmath $7.95$ & $7.49$ & $m$\\
\hline
min\_dist\_to\_people & $0.40$ & \boldmath $0.90$ & $0.55$ & $m$\\
\hline
intimate\_space\_intrusions & $3.66$ & \boldmath $0$ & \boldmath $0$ & $\%$\\
\hline
personal\_space\_intrusions & $19.7$ & \boldmath $5.8$ & $14.7$ & $\%$\\
\hline
social+\_space\_intrusions & $17.0$ & \boldmath $26.7$ & $21.4$ & $\%$ \\
\hline
completed & $100$ & $100$ & $100$ & $\%$ \\
\hline
min\_dist\_to\_target & $0.18$ & $0.19$ & \boldmath $0.11$ & $m$ \\
\hline
final\_dist\_to\_target & $0.18$ & $0.19$ & \boldmath $0.11$ & $m$\\
\hline
robot\_on\_person\_collisions & $0$ & $0$ & $0$ & - \\
\hline
person\_on\_robot\_collisions & $0$ & $0$ & $0$ & - \\
\hline
time\_not\_moving & \boldmath $0.00$ & $1.04$ & $1.78$ & $s$\\
\hline
avg\_robot\_linear\_speed & \boldmath $0.51$ & $0.51$ & $0.38$ & $m/s$\\
\hline
avg\_robot\_angular\_speed & \boldmath $0.02$ & $0.09$ & $0.09$ & $rad/s$\\
\hline
avg\_robot\_acceleration & \boldmath $0.05$ & $0.05$ & $0.08$ & $m/s^2$\\
\hline
avg\_robot\_jerk & $0.08$ & \boldmath $0.06$ & $0.08$ & $m/s^3$\\
\hline
avg\_pedestrian\_velocity & \boldmath $0.47$ & $0.44$ & $0.34$ & $m/s$\\
\hline
avg\_closest\_pedestrian\_velocity & \boldmath $0.57$ & $0.51$ & $0.39$ & $m/s$\\
\hline
social\_force\_on\_agents & $0.52$ & $0.97$ & \boldmath $0.49$ & $m/s^2$\\
\hline
social\_force\_on\_robot & $0.52$ & $0.92$ & \boldmath $0.49$ & $m/s^2$\\
\hline
obstacle\_force\_on\_agents & $8.14$ & \boldmath $0.13$ & $0.81$ & $m/s^2$\\
\hline
obstacle\_force\_on\_robot & $0.00$ & $0.00$ & $0.00$ & $m/s^2$ \\
\hline
social\_work & $1.04$ & $1.89$ & \boldmath $0.98$ & $m/s^2$\\
\hline
\end{tabular}
\end{center}
\end{table}

We can draw some conclusions from these results. First, the \textit{DWB} planner does not consider any social component, so its objective is to reach the goal as soon as possible without colliding. Therefore it obtains the best performance in metrics like time to reach the goal, path length or the average robot velocities. 

  \begin{figure}[!t] 
      \centering
      \includegraphics[scale=0.55]{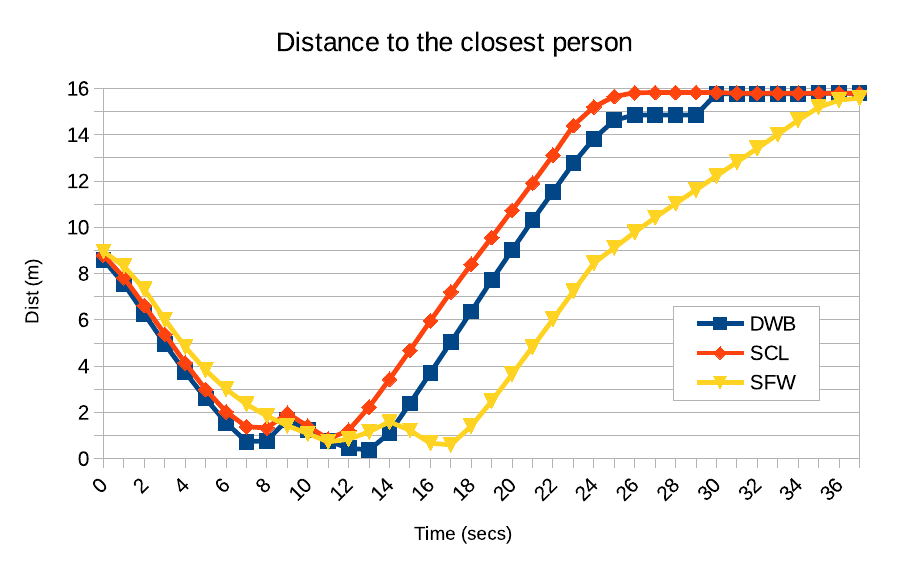}
      \caption{Distance to the closest person for one trajectories of the \texttt{Passing} scenario.}
      \label{fig:dist_closest_passing1}
   \end{figure}

Secondly, the \textit{SCL} adds an extra cost to the navigation costmap around the people. That provokes the robot to keep more distance from people. In the \texttt{Passing} scenario, that is reflected in metrics like the average distance to the closest person, minimum distance to people and the intrusions in the intimate and personal spaces. However, we observe in the \texttt{Crossing} scenario that several people may come close to the robot at the same time, which provokes high costs all around the robot. In this situation, with high costs for most of the potential trajectories, the robot decides to move towards the goal, leading to higher intimate space intrusions, lower average distance to the closest person, and higher social force on agents than in the \texttt{Passing} scenario. 

Finally, the \textit{SFW} planner includes human-aware constraints. Specifically, it uses the social work as a term in its trajectory scoring function. That leads the robot to better preserve the comfort of surrounding people. It reaches the best performance in metrics like social work and social force on agents as could be expected. It also reaches the best results in the intrusions in the intimate space of people, and it obtains good results in the minimum distance to people or the average distance to the closest person. It also spends more time stopped (probably yielding).

\begin{table}[t]
\caption{Metrics for the \texttt{Crossing} scenario}
\label{tab:crossing}
\begin{center}
\begin{tabular}{|c||c|c|c|c|}
\hline 
\texttt{Crossing} & \textit{DWB} & \textit{SCL} & \textit{SFW} & units \\
\hline \hline
time\_to\_reach\_goal & \boldmath $10.1$ & $14.6$ & $14.0$ & $s$\\
\hline
path\_length & \boldmath $4.9$ & $5.7$ & \boldmath $4.9$ & $m$\\
\hline
cumulative\_heading\_changes & $2.15$ & $6.68$ & \boldmath  $2.03$ & $rad$\\
\hline
avg\_dist\_to\_closest\_person & $1.45$ & $1.37$ & \boldmath $1.51$ & $m$\\
\hline
min\_dist\_to\_people & $0.23$ & $0.21$ & \boldmath $0.49$ & $m$\\
\hline
intimate\_space\_intrusions & $13.7$ & $14.6$ & \boldmath $1.6$ & $\%$\\
\hline
personal\_space\_intrusions & $28.5$ & \boldmath $25.2$ & $44.9$ & $\%$\\
\hline
social+\_space\_intrusions & $57.9$ & \boldmath $60.2$ & $53.4$ & $\%$ \\
\hline
completed & $100$ & $100$ & $100$ & $\%$ \\
\hline
min\_dist\_to\_target & $0.19$ & $0.22$ & \boldmath $0.13$ & $m$ \\
\hline
final\_dist\_to\_target & $0.19$ & $0.22$ & \boldmath $0.13$ & $m$\\
\hline
robot\_on\_person\_collisions & $0$ & $0$ & $0$ & - \\
\hline
person\_on\_robot\_collisions & $0$ & $0$ & $0$ & - \\
\hline
time\_not\_moving & \boldmath $0.00$ & \boldmath $0.00$ & $0.38$ & $s$\\
\hline
avg\_robot\_linear\_speed & \boldmath $0.35$ & $0.31$ & $0.28$ & $m/s$\\
\hline
avg\_robot\_angular\_speed & $0.44$ & $0.55$ & \boldmath $0.25$ & $rad/s$\\
\hline
avg\_robot\_acceleration & \boldmath $0.09$ & $0.12$ & $0.14$ & $m/s^2$\\
\hline
avg\_robot\_jerk & \boldmath  $0.10$ & $0.13$ & $0.26$ & $m/s^3$\\
\hline
avg\_pedestrian\_velocity & \boldmath $0.46$ & $0.36$ & $0.36$ & $m/s$\\
\hline
avg\_closest\_pedestrian\_velocity & $0.27$ & \boldmath $0.31$ & $0.23$ & $m/s$\\
\hline
social\_force\_on\_agents & $0.67$ & $1.10$ & \boldmath $0.44$ & $m/s^2$\\
\hline
social\_force\_on\_robot & $0.67$ & $1.10$ & \boldmath $0.44$ & $m/s^2$\\
\hline
obstacle\_force\_on\_agents & \boldmath  $2.08$ & $6.07$ & \boldmath  $2.08$ & $m/s^2$\\
\hline
obstacle\_force\_on\_robot & $0.00$ & $0.00$ & $0.00$ & $m/s^2$ \\
\hline
social\_work & $1.33$ & $2.20$ & \boldmath $0.88$ & $m/s^2$\\
\hline
\end{tabular}
\end{center}
\end{table}

    \begin{figure}[!t] 
      \centering
      \includegraphics[scale=0.57]{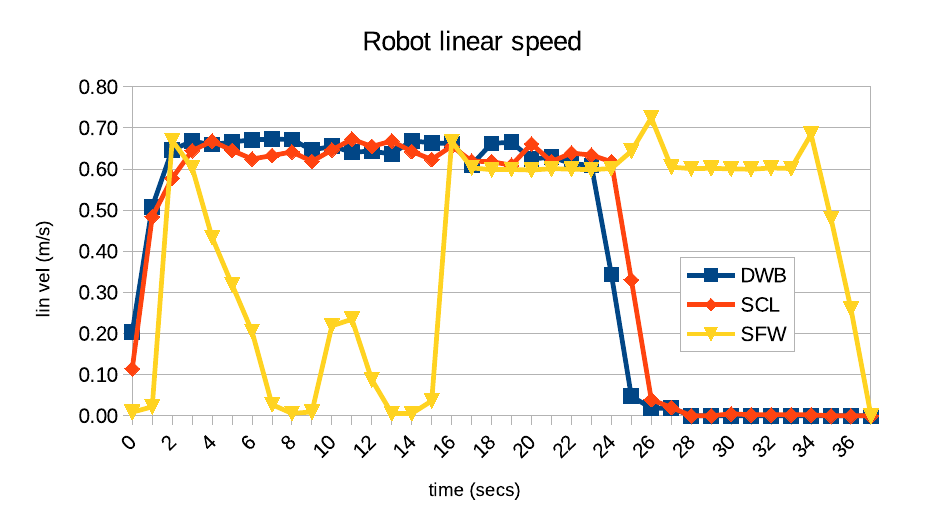}
      \caption{Robot linear velocity for one trajectories of the \texttt{Passing} scenario.}
      \label{fig:lin_vel_passing1}
   \end{figure}

\begin{table*}[h]
\caption{Some metrics for the planners in the \texttt{Combined} scenario with people showing different behaviors. The order of the shown values is \textit{DWB} planner, \textit{SCL} planner and \textit{SFW} planner.}
\label{tab:combined}
\begin{center}
\begin{tabular}{|c||c|c|c|}
\hline 
 \textit{DWB} / \textit{SCL} / \textit{SFW} & Regular & Curious & Threatening \\
\hline \hline
min\_dist\_to\_people & $0.96$ / $\mathbf{1.07}$ / $1.05$ & $0.77$ / $\mathbf{0.80}$ / $0.75$ & $\mathbf{0.21}$ / $0.20$ / $0.15$ \\
\hline
avg\_dist\_to\_closest\_person & $11.42$ / $11.31$ / $\mathbf{12.85}$ & $3.56$ / $3.99$ / $\mathbf{8.05}$ & $\mathbf{3.55}$ / $3.54$ / $3.29$ \\
\hline
intimate\_space\_intrusions & $0.00$ / $0.00$ / $0.00$ & $0.00$ / $0.00$ / $0$ & $52.84$ / $\mathbf{45.24}$ / $58.46$ \\
\hline
personal\_space\_intrusions & $4.27$ / $4.76$ / $\mathbf{1.54}$ & $\mathbf{10.03}$ / $11.90$ / $13.85$ & $14.56$ / $21.43$ / $\mathbf{4.62}$ \\
\hline
avg\_pedestrian\_velocity & $\mathbf{0.22}$ / $0.21$ / $0.13$ & $\mathbf{0.59}$ / $0.56$ / $0.46$ & $0.66$ / $\mathbf{0.67}$ / $0.60$ \\
\hline
social\_force\_on\_robot & $0.00$ / $0.00$ / $0.00$ & $\mathbf{0.07}$ / $0.08$ / $0.30$ & $3.23$ / $\mathbf{1.91}$ / $2.91$ \\
\hline
social\_work & $0.00$ / $0.00$ / $0.00$ & $\mathbf{0.14}$ / $0.15$ / $0.60$ & $6.46$ / $\mathbf{3.83}$ / $5.83$ \\
\hline
\end{tabular}
\end{center}
\end{table*}

The \textit{HuNavSim} also provides the values of the metrics along the robot trajectory. For instance, Fig. \ref{fig:dist_closest_passing1} shows the distance to the closest person for each time step of one of the trajectories performed of the \texttt{Passing} scenario. We can observe the moments in which the robot passes the two pedestrians, and how the \textit{DWB} is the planner which chooses to pass closer to them. The same situation is shown in Fig. \ref{fig:lin_vel_passing1} for the robot linear speed. In this case, we clearly see how the \textit{SFW} planner decides to almost stop when very close to people in order to not disturb them. 

In summary, planners including only distance constrains (like \textit{SCL}) should expect to obtain good results in distance-based metrics. More advanced planners which also predict future people positions regarding velocities and directions (like \textit{SFW}) should obtain better results in metrics based on social forces and social work, since they take into account people distances, velocities and directions. The planners can be better characterized thanks to having a broad set of metrics.

\subsubsection{Results for different navigation behaviors}

In a realistic scenario, not only do people walk at different velocities and maintain different distances, but they also react very differently to the presence of an autonomous mobile robot. Some people will treat the robot like another pedestrian while others might feel curiosity, mistrust or even show harassing behavior. In most cases, these possible reactions are neither taken into account while developing robotic navigation algorithms nor are they considered in the current pedestrian simulators.

We evaluate here the results obtained according to the different behaviors of the three human agents in the \texttt{Combined} scenario. Table \ref{tab:combined} shows the results of the three planners for some interesting metrics related to the human behavior. For the three planners some general conclusions can be extracted: in the case of the ``regular'' human, we obtain the expected metrics as above. However, as we observe, the ``curious'' and ``threatening'' agents provoke totally different metrics since they abandon their regular navigation to approach the robot for a while. We also see the differences between a ``curious'' person and a ``threatening'' person. The latter is more intrusive, since the human agent tries to block the robot path and moves very close to the robot. In that situation, the \textit{SFW} planner prefers the robot to spend more time almost fully stopped than the other planners to prevent disturbances and possible collisions. The robot navigation algorithms must be very careful since the risk of collision is high. 

These results emphasize the complexity of performing robotic navigation experiments in real and uncontrolled scenarios. Additionally, computing significant metrics without considering the different reactions of the pedestrians can lead to wrong or unclear results. \textit{HuNavSim} can help to develop and to benchmark navigation algorithms that consider these possible reactions.

\section{Conclusions and Future Work}

We have introduced a new open-source software library used to simulate human navigation behaviors, \textit{HuNavSim}. The tool, programmed under the new ROS 2 framework, can be employed to control the human agents of different general robotics simulators. A wrapper for the use of the tool with the well-known Gazebo simulator is also introduced. 

Moreover, it presents other novelties, such as a set of individualized human navigation behaviors directed by behavior trees. These behaviors are easily configurable and expandable due to the use of behavior trees. The tool also includes an extensive, flexible and extensible compilation of evaluation metrics for robot human-aware navigation taken from the literature.   

The provided comparison of three different planners demonstrates that the tool is useful for the development of navigation algorithm and for the evaluation of them. The addition of different reactions to the presence of the robot also presents a novelty that helps to simulate more realistic navigation situations in spaces shared with humans.

Future work includes the extension of the set of metrics. 
The metrics from the \textit{Crowdbot} \cite{crowdbot_ICRA21} and others indicated in the work of Gao \emph{et al}. \cite{gao_frontiers22}, will be studied and included. The improvement of the body animations accompanying the agents movements in the Gazebo wrapper is being considered just as the development of wrappers for other simulators like Webots or Isaac Sim. Moreover, the augmentation of the set of individual navigation behaviors will be studied, as well as adding other human-navigation models besides the Social Force Model. Finally, on a technical level, we will work to update the software to use the ROS 2 Humble distribution (under development currently) and the latest version of the Behavior Tree library (v4).     




\balance
\bibliographystyle{IEEEtran} 
\bibliography{hunav}

\end{document}